\documentclass[sigconf]{acmart}
\usepackage{soul, color, xcolor}
\usepackage{multirow}
\usepackage{makecell}
\usepackage[table]{xcolor}
\usepackage[most]{tcolorbox}
\usepackage{utfsym}
\usepackage{color}
\usepackage{lipsum}
\usepackage{pifont}

\usepackage{graphicx}
\usepackage[export]{adjustbox}
\AtBeginDocument{%
  }

\copyrightyear{2025}
\acmYear{2025}
\acmConference[KDD '25]{Proceedings of the 31st ACM SIGKDD Conference on Knowledge Discovery and Data Mining V.2}{August 3--7, 2025}{Toronto, ON, Canada}
\acmBooktitle{Proceedings of the 31st ACM SIGKDD Conference on Knowledge Discovery and Data Mining V.2 (KDD '25), August 3--7, 2025, Toronto, ON, Canada}
\acmDOI{}
\acmISBN{}

\begin{document}


\title{QA-Dragon: Query-Aware Dynamic RAG System for Knowledge-Intensive Visual Question Answering}


\author{Zhuohang Jiang}
\authornote{Authors contributed equally to this research.}
\affiliation{%
  \institution{The Hong Kong Polytechnic University}
   \city{Hong Kong SAR}
  \country{China}
}
\email{zhuohang.jiang@connect.polyu.hk}

\author{Pangjing Wu}
\authornotemark[1]
\affiliation{%
  \institution{The Hong Kong Polytechnic University}
   \city{Hong Kong SAR}
  \country{China}
}
\email{pang-jing.wu@connect.polyu.hk}

\author{Xu Yuan}
\authornotemark[1]
\affiliation{%
  \institution{The Hong Kong Polytechnic University}
   \city{Hong Kong SAR}
  \country{China}
}
\email{xander.yuan@connect.polyu.hk}

\author{Wenqi Fan}
\authornote{Prof. Qing Li and Prof. Wenqi Fan are the advisors of the team.}
\affiliation{%
  \institution{The Hong Kong Polytechnic University}
   \city{Hong Kong SAR}
  \country{China}
}
\email{wenqi.fan@polyu.edu.hk}

\author{Qing Li}
\authornotemark[2]
\affiliation{%
  \institution{The Hong Kong Polytechnic University}
   \city{Hong Kong SAR}
  \country{China}
}
\email{csqli@comp.polyu.edu.hk}

\renewcommand{\shortauthors}{Zhuohang Jiang et al.}
\begin{abstract}
Retrieval-Augmented Generation (RAG) has been introduced to mitigate hallucinations in Multimodal Large Language Models (MLLMs) by incorporating external knowledge into the generation process, and it has become a widely adopted approach for knowledge-intensive Visual Question Answering (VQA).
However, existing RAG methods typically retrieve from either text or images in isolation, limiting their ability to address complex queries that require multi-hop reasoning or up-to-date factual knowledge.
To address this limitation, we propose \textbf{QA-Dragon}, a \textbf{\underline{Q}}uery-\textbf{\underline{A}}ware \textbf{\underline{D}}ynamic \textbf{\underline{RAG}} System for Kn\textbf{\underline{o}}wledge-I\textbf{\underline{n}}tensive VQA.
Specifically, QA-Dragon introduces a domain router to identify the query’s subject domain for domain-specific reasoning, along with a search router that dynamically selects optimal retrieval strategies.
By orchestrating both text and image search agents in a hybrid setup, our system supports multimodal, multi-turn, and multi-hop reasoning, enabling it to tackle complex VQA tasks effectively.
We evaluate our QA-Dragon on the Meta CRAG-MM Challenge at KDD Cup 2025, where it significantly enhances the reasoning performance of base models under challenging scenarios.
Our framework achieves substantial improvements in both answer accuracy and knowledge overlap scores, outperforming baselines by 5.06\% on the single-source task, 6.35\% on the multi-source task, and  5.03\% on the multi-turn task. The source code for our system is released in \href{https://github.com/jzzzzh/QA-Dragon}{https://github.com/jzzzzh/QA-Dragon}.

\end{abstract}

\ccsdesc[500]{Information systems~Information retrieval}

\keywords{Retrieval-Augmented Generation, Multimodal Large Language Model, Visual Question Answering}



\maketitle

\section{Introduction}
Powered by advanced Large Language Models (LLMs)~\cite{ouyang2022training,chiang2023vicuna,touvron2023llama,chowdhery2023palm} and sophisticated visual perception modules, Multimodal Large Language Models (MLLMs)~\cite{li2023blip,liu2023visual,lin2024vila,wang2409qwen2} have demonstrated strong understanding and reasoning capabilities across a wide range of vision-language tasks, such as Visual Question Answering (VQA).
Despite these advancements, MLLMs still face significant challenges when addressing queries that require long-tail knowledge and multi-hop reasoning, often generating hallucinated or inaccurate responses~\cite{qiu2024snapntell,cocchi2025augmenting}.
These issues stem from the scarcity of relevant knowledge in MLLMs’ training corpus and the inherent difficulty of memorizing low-frequency facts~\cite{chen2023can}. 
Retrieval-Augmented Generation (RAG)~\cite{fan2024survey} has recently emerged as a promising solution, promoting MLLMs by incorporating external information to complement their internal knowledge.
However, multimodal RAG (MM-RAG) still faces significant challenges, such as interpreting complex queries, selecting appropriate retrieval tools, refining relevant information, and enabling effective multi-turn interactions.

To address these challenges, we introduce a Query-Aware Dynamic RAG System for Knowledge-Intensive VQA (\textbf{QA-Dragon}), specifically designed to address cross-domain, knowledge-based, and multi-hop reasoning VQA tasks. QA-Dragon incorporates three specialized reasoning branches and a series of modular components, including a Pre-Answer Module, Search Router, Tool Router, Image \& Text Retrieval Agents, a Multimodal Reranker, and a Post-Answer Module. These components combine to dynamically select the optimal retrieval strategy based on domain-specific query characteristics. Specifically, the Search Router interprets the intent of a given query. It dispatches it to the appropriate retrieval pathway, while the Tool Router further refines the execution by selecting between image-based and text-based retrieval agents according to the query modality. To ensure the relevance and quality of the retrieved information, the multimodal reranker uses a coarse-to-fine refinement process that grounds final responses in high-quality evidence.

By supporting multimodal, multi-turn, and multi-hop reasoning, QA-Dragon can address the complexity of real-world VQA scenarios. Empirical results demonstrate that our framework yields substantial improvements in both answer accuracy and knowledge overlap, outperforming strong baselines by 5.06\% on the single-source task, 6.35\% on the multi-source task, and 5.03\% on the multi-turn task.

\section{Competition Description}
To highlight the challenges of real-world VQA tasks, Meta and AIcrowd launched the Meta CRAG‑MM Challenge 2025~\cite{crag-mm-2025}, an official KDD Cup competition focused on comprehensive RAG for multimodal, multi-turn question answering (QA) over images captured in the wild. 
This benchmark combines egocentric photos from Ray‑Ban Meta smart glasses with factual QA pairs, a mock image-based knowledge base, web search APIs, and rigorous truthfulness scoring, offering the first large-scale testbed for end-to-end MM-RAG systems.

\subsection{Dataset}
The \textbf{CRAG-MM} release comprises three coordinated resources: an \textit{Image Set}, a \textit{QA Set}, and \textit{Retrieval Database}.

\begin{itemize}
    \item \textbf{Image Set} contains 5,000 RGB images, of which 3,000 are first-person ``egocentric'' shots from smart-glasses, while the rest are ordinary photos scraped from the open web. These images span 14 topical domains (e.g., Books, Food, Shopping, Vehicles) and intentionally include low-quality or cluttered views that emulate real-world wearable scenarios.

    \item \textbf{QA Set} includes two complementary partitions:
    \begin{enumerate}
        \item \textbf{Single-turn}: More than 3.88k independent QA pairs (1.94k validation + 1.94k public-test).
        \item \textbf{Multi-turn}: 1{,}173 dialog sessions (586 validation + 587 public-test) comprising 2-6 interleaved questions per image.
    \end{enumerate}

    \item \textbf{Retrieval Database} is composed of two controlled sources designed to equalize access across teams:
    \begin{enumerate}
        \item An \textit{image-KG mock API} that returns visually similar images and structured metadata (title, brand, price, etc.) keyed on the query image.
        \item A \textit{textual web-search mock API} that yields URL, title, snippet, timestamp, and full HTML for up to 50 pages per query, interleaving hard negatives to mimic real-world noise.
    \end{enumerate}
\end{itemize}

Together, these ingredients create a realistic yet reproducible sandbox for studying hallucination-free answer generation in wearable contexts.

\subsection{Search Engine}

To support faithful answer generation grounded in external knowledge, the CRAG-MM Challenge provides two unified, Python-based mock retrieval APIs—one for images and one for textual web content—forming the backbone of the RAG pipeline. These APIs simulate realistic retrieval conditions while ensuring a level playing field across participants.

\subsubsection{Image Search API} This API enables the retrieval of structured entity-level metadata from visually similar images. It uses CLIP-based image encoders~\footnote{https://huggingface.co/openai/clip-vit-large-patch14} to embed both the query image and database images, returning top-$k$ results based on cosine similarity. Each result includes a URL and associated structured metadata such as \textit{title}, \textit{brand}, \textit{price}, and \textit{description}, mimicking a knowledge graph interface grounded in vision. This API is the only retrieval source available in Task~1.

\subsubsection{Web Search API} A textual search interface is added in Tasks~2 and~3. This API indexes pre-fetched web pages using ChromaDB and supports semantic search over HTML content. Given a text query, it returns up to 50 web pages, each with a \textit{URL}, \textit{title}, \textit{snippet}, and timestamp. Relevance is computed via sentence-transformer embeddings (\emph{e.g.}, \textit{bge-large-en-v1.5}~\cite{bge_embedding}) using cosine similarity. Hard negatives are interleaved to reflect real-world retrieval noise.

\subsection{Tasks}

The competition defines three progressively harder tasks, including  \textbf{Single-source Augmentation}, \textbf{Multi-source Augmentation} and \textbf{Multi-turn QA} tasks.

\begin{itemize}
    \item \textbf{Task \#1 - Single-source Augmentation}: Given \textit{only} the image-KG API, the system must retrieve structured facts linked to the image and produce a grounded answer. This evaluates core visual recognition, query reformulation, and KG grounding.

    \item \textbf{Task \#2 - Multi-source Augmentation}: Web-search API results are added, requiring the system to fuse heterogeneous evidence, filter noise, and justify answers drawn from both the image KG and the web.

    \item \textbf{Task \#3 - Multi-turn QA}: Dialog sessions of 2-6 turns probe contextual understanding, answer consistency, and the ability to decide when the current image is still relevant versus when text-only reasoning suffices. Systems must respect a strict 10~s per-turn latency and a 30~s total response budget.
\end{itemize}
\begin{figure*}
    \centering
    \includegraphics[width=\linewidth]{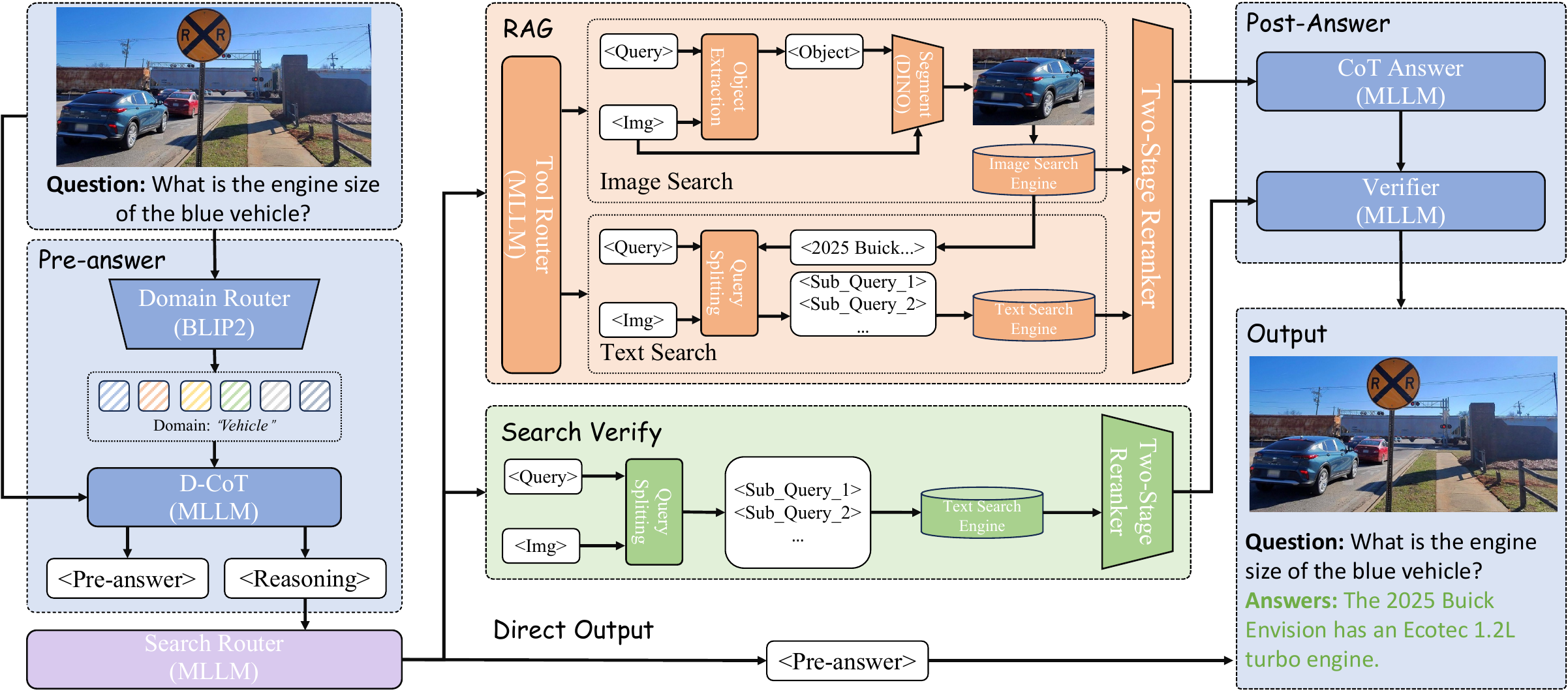}
    \caption{An overview of the QA-Dragon framework. Given a multimodal query–image pair, the system first processes the input through a \textit{Domain Router} and D-CoT to produce a draft answer and reasoning trace. A \textit{Search Router} then classifies the query into one of three branches: \textit{a) Direct Output}, which emits the answer immediately; \textit{b) Search Verify}, which retrieves external evidence for verification; and \textit{c) RAG}, which performs retrieval via tool routing, multimodal search, and reranking. The resulting evidence is passed to the \textit{Post-Answer Module} for final verification and answer refinement.}
    \label{fig:pipeline}
\end{figure*}

\section{Methodology}
To address the challenge of grounded and trustworthy answering for real-world multimodal queries, we propose \textbf{QA-Dragon}, a framework that integrates domain-aware reasoning, adaptive retrieval, and trustworthiness verification. QA-Dragon decomposes the problem into three branches with multiple processes: 1) a \textbf{Pre-Answer Module}, which performs domain classification and generates an initial reasoning trace and answer  using a domain-specific Chain-of-Thought (D-CoT) agent; 2) a \textbf{Search Router}, which inspects the reasoning trace to determine whether additional external evidence is required, and if so, selects between retrieval-augmented generation or answer verification; 3) a \textbf{Tool Router}, which decides whether to invoke a 4) \textbf{Image Search Agent} or a 5) \textbf{Text Search Agent} to search for useful information, 6) a \textbf{Coarse-to-Fine Multimodal Reranker}, which selects the most relevant information to augment answers, and 7) a \textbf{Post-Answer Module}, which consolidates retrieved evidence with the initial hypothesis to generate a final, verifiable response. By explicitly modeling domain context and evidence needs, QA-Dragon improves factual consistency, supports low-latency fallback through direct output, and mitigates hallucination via adaptive verification and reranking strategies.

\subsection{Pre-Answer Module}

\subsubsection{Domain-Router}
Real-world multimodal queries are highly diverse, spanning topics such as animals, food, vehicles, and complex data visualizations. Treating all queries uniformly risks suboptimal reasoning due to domain mismatch and lack of specialized context. To address this, we introduce a domain router, which identifies the semantic domain of each query–image pair and enables domain-specific reasoning strategies tailored to the multimodal queries.

The domain router predicts the semantic domain of the input $(x,q)$ pair, where $x$ is an image and $q$ is the corresponding textual query, allowing the system to invoke specialized reasoning agents with in-domain examples and tailored prompts. Specifically, we leverage BLIP-2~\cite{li2023blip}, an efficient vision–language model, to bridge image understanding and language reasoning through a lightweight query transformer. It can jointly encode visual and textual signals, enabling robust multimodal classification. 

The BLIP-2 is finetuned using the domain annotations available in the competition dataset. The predicted domain label $d$ is then used to dispatch the query to the corresponding D-CoT process, which performs in-domain reasoning under customized prompting templates.

\subsubsection{D-CoT}
While LLMs exhibit strong generalization, they often lack awareness of their knowledge boundaries~\cite{qiu2024snapntell,cocchi2025augmenting}. 
To mitigate this, we introduce the D-CoT, which performs a structured pre-answer reasoning process to identify what the model confidently knows and where it lacks sufficient information. 
This early introspective step enables the system to decide whether further evidence retrieval is needed.

The D-CoT module prompts the MLLM to generate step-by-step and domain-aware reasoning grounded in visual content and the user query. 
It is accomplished by composing a domain-specific prompt with a small number of curated in-domain few-shot examples, forming an input sequence. Specifically, we prompt MLLM to generate a provisional answer and an explicit reasoning trace. To ensure domain-relevant and interpretable reasoning, each prompt includes explicit behavioral constraints. The model must 1) identify the exact object referenced by the query, 2) reason over image and contextual clues step-by-step, and 3) explicitly signal uncertainty when necessary information is missing, as illustrated in the Appendix.

This instruction reveals the model's internal logic and provides a transparent basis for downstream routing decisions. The resulting reasoning trace is then passed to the search router, determining whether the provisional answer is sufficient or if further retrieval or verification steps should be invoked.

\newcolumntype{C}[1]{>{\centering\arraybackslash}m{#1}}
\begin{table*}[ht]
\caption{Execution Path Selection based on CoT Output and Router Cues.}
\label{tab:routing}
\centering
\begin{tabular}{|c|C{3.2cm}|C{4.2cm}|C{2.6cm}|C{3.8cm}|}
\hline
\textbf{Branch} & \textbf{Trigger} & \textbf{Processing Modules} & \textbf{Image} & \textbf{Query} \\
\hline
Direct Output & D-CoT succeeds, and the query is classified as self-contained (\emph{e.g.}, arithmetic, OCR) 
& -- 
& \hspace{0mm}\raisebox{-1mm}{\includegraphics[width=2.5cm]{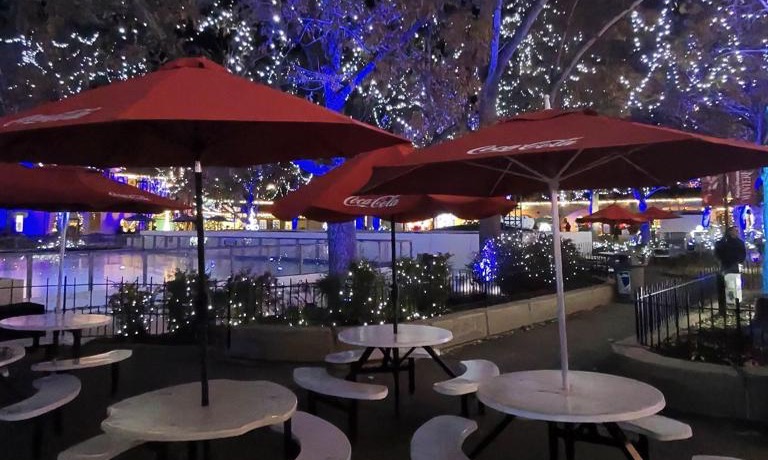}}
& \textit{“What is written on these umbrellas?”} \\
\hline
Search Verify 
& D-CoT succeeds, but external evidence is recommended for verification 
& Text Toolchain $\rightarrow$ Fusion \& RAG Engine $\rightarrow$ Verifier 
& \hspace{0mm}\raisebox{-1mm}{\includegraphics[width=2.5cm]{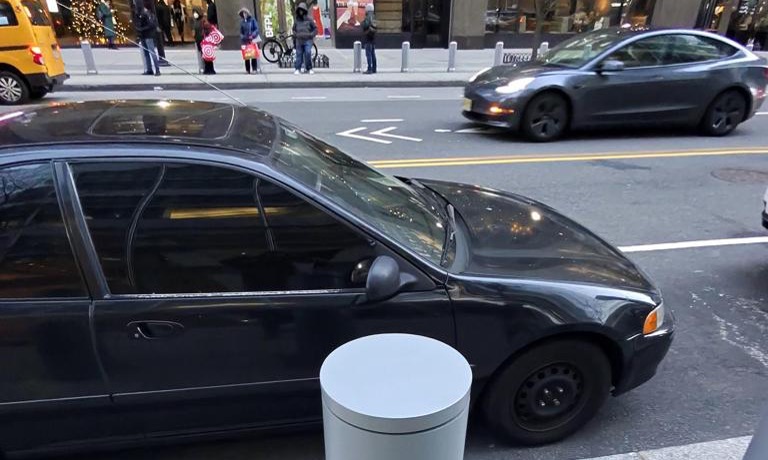}}
& \textit{“In which year did the car on the right begin production?”} \\
\hline
RAG-Augment 
& D-CoT fails (\emph{e.g.}, “I don't know”) or Router detects open-world cues (\emph{e.g.}, entities, numbers)
& Visual Toolchain $\rightarrow$ Text Toolchain $\rightarrow$ Fusion \& RAG Engine $\rightarrow$ Verifier
& \hspace{0mm}\raisebox{-1mm}{\includegraphics[width=2.5cm]{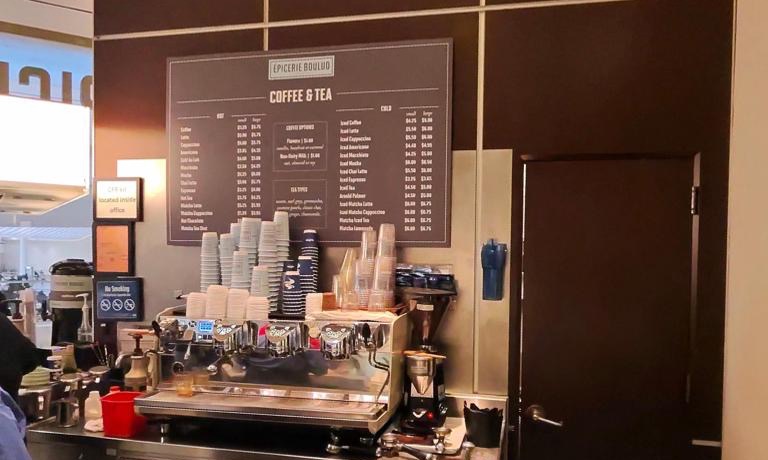}}
& \textit{“Who founded this cafe?”} \\
\hline
\end{tabular}
\end{table*}

\subsection{Search Router}
Not all queries require the same depth of post-processing. Some queries can be resolved directly based on image-grounded information, such as extracting OCR text or calculating visual formulas. Others may require external evidence or verification to ensure factual accuracy. To make these decisions adaptively, we introduce the Search Router, a key component that selects the most suitable execution path based on the pre-answer and reasoning trace from the D-CoT.

The search router is to determine whether to directly output the preliminary answer, perform factual verification using external sources, or invoke a retrieval-augmented generation process to synthesize a new, grounded response. This routing mechanism allows the system to balance efficiency and reliability, ensuring that each query receives just the right amount of computational effort for trustworthy results.

The router takes as input the original query and image, along with the draft answer, reasoning trace, and predicted domain from the D-CoT module. Instead of relying solely on the language model’s self-assessment, the router is implemented as a lightweight classifier trained on features derived from the reasoning trace and answer content. These features include:

\begin{itemize}
    \item \textbf{Answerability Flags:} Whether the reasoning contains uncertain phrases like \textit{``I don't know''} or falls into known failure patterns.
    \item \textbf{Answer Task Heuristics:} Whether the answer is a number, an OCR string, or a named object, indicating that the information is directly observable from the image.
    \item \textbf{Uncertainty Patterns:} Whether the reasoning exhibits speculative or fallback logic that may require further verification.
\end{itemize}

Based on these indicators, the router assigns each query to one of three execution paths: \textbf{Direct Output}, when the answer can be confidently derived from the image alone; Search \textbf{Verify}, when the draft answer lacks certainty and requires evidence checking; and \textbf{RAG}, when the query depends on external knowledge not present in the image or reasoning trace.

\subsection{Tool Router}
To efficiently retrieve missing information, our framework must determine not only whether to search, but also what kind of search to perform. Queries differ in the type of evidence required: some lack object identity (\emph{e.g.,} \textit{``What model is this?''}), while others require factual attributes not visible in the image (\emph{e.g.,}  \textit{``What is the price of this car?''}). Motivated by this need for precision in retrieval, we introduce the tool router, a lightweight decision module that selects the appropriate retrieval modality, including \textbf{image search}, \textbf{text search}, \textbf{both}, or \textbf{neither}, based on the reasoning trace obtained from the pre-answer module. 

Operating after the D-CoT module, the tool router evaluates the reasoning trace to assess the model’s knowledge boundaries. If the object mentioned in the query has not been identified with a specific name, it will initiate an image search to retrieve visually similar items and infer the object's identity. In contrast, if the LLM has already identified the object but lacks necessary facts that are not inferable from the image, such as specifications, statistics, or pricing, it triggers a text search to collect relevant information from the web. Queries that involve analytical reasoning tasks, such as mathematical calculations or language translation, are typically self-contained and do not require retrieval. The key decision prompt is illustrated in Appendix~\ref{sec:prompt}.

\subsection{Image Search Agent}
The image search agent is responsible for grounding visual entities relevant to the query. It combines \textbf{Multimodal Object Extraction}, \textbf{Segmentation}, \textbf{Multi-image Search Engine}, and \textbf{Entity Selection Module} to prepare visual evidence for answer augmentation.

\subsubsection{Multimodal Object Extraction.} 
To identify the visual entities relevant to a user query, we employ the LLM to extract a set of grounded object candidates from the input image and question. 

Given an image and its associated query, the model predicts a list of salient and visually present objects that are likely to be referenced or needed for downstream reasoning. Each extracted object is described using concise, coarse-grained terms, such as ``car,'' ``book,'' or ``brand'', and is constrained to represent tangible items only, explicitly excluding abstract concepts and actions. To ensure generalization across domains, the object names are normalized into category-level descriptions and are limited to short, interpretable phrases of up to three words. This process yields a structured object set that supports subsequent steps in query grounding, answer generation, and retrieval conditioning.

From the extracted candidates, a secondary selection step identifies the most relevant object to the query. This object is chosen based on its visual presence, contextual alignment with the query, and positional cues in the image. When multiple similar instances exist, the model appends a distinguishing attribute (\emph{e.g.}, “red car,” “white car”) to ensure precise reference. The result is a single object label that serves as the anchor for following image segmentation.

\subsubsection{Segmentation.}
In the VQA process, users' questions often focus on a specific object in an image rather than the entire image. To achieve more granular retrieval, we introduce a refined search mechanism based on image segmentation. Specifically, we extract key object words from the image during the multimodal object extraction stage, use grounding DINO~\cite{liu2023grounding} to locate and crop these targets, and generate corresponding local image regions for subsequent retrieval. For images containing multiple objects with the same name, our system also supports the simultaneous extraction and processing of multiple related regions to more comprehensively cover the user's query intent.

\subsubsection{Multi-Image Search Engine.}
To meet the needs of single-image or multi-image retrieval, we designed and implemented a customised image retrieval engine that can automatically adapt to multiple image regions generated during the image segmentation stage. The engine performs searches based on the \textit{"crag-mm-2025/image-search-index-validation"}\cite{crag-mm-2025} index and uses the \textit{"clip-vit-large-patch14-336"}\cite{radford2021learning} model to calculate the semantic embedding similarity between the query and candidate image results. Finally, we fuse multiple retrieval results and reorder them based on similarity scores to obtain image content more closely related to the query semantics, thereby providing high-quality visual support for multimodal QA.  

\subsubsection{Entity Selection.}
When external image retrieval is performed, it is essential to ensure that the retrieved candidates visually correspond to the actual object depicted in the image. Without this verification, downstream reasoning may rely on mismatched entities, leading to hallucinated or irrelevant answers. To address this challenge, we introduce an entity selection step, which verifies whether any retrieved entity is visually present and consistent with the object shown in the query image.

\subsection{Text Search Agent}
The text search agent is responsible for retrieving external textual evidence relevant to a given query–image pair. It is composed of three sequential components: the \textbf{Query Rephrase}, the \textbf{Fusion Search} and the \textbf{Text Search Engine} to handle multi-hop reasoning, entity disambiguation, and weakly grounded queries.

\subsubsection{Query Rephrasing Module.}
Since search engines require exact matches, it is difficult to handle multi-hop or ambiguous queries. Therefore, we need to break down multi-hop queries into multiple clear sub-queries and replace any references or ambiguous content to improve search results.

Specifically, we first structure and break down the complex original queries. Using the original query, the reasoning chain of thought, and image information extracted by the visual tool chain, we construct multiple more specific sub-questions, each of which typically corresponds to an inference step in the original question or information related to a specific image entity. To further enhance the clarity and retrieval adaptability of these sub-questions, we introduce a sub-question enhancement module to semantically clarify and complete each generated sub-queries, particularly addressing parts with unclear referents or lacking context.

\subsubsection{Fusion Search}
The fusion search serves as a bridge between the image and text retrieval pipelines by leveraging grounded object information extracted during the image search stage to improve the quality and precision of text-based evidence retrieval. This is particularly useful when the original user query lacks a specific object identity or when disambiguation is required to generate a meaningful web search query.

After identifying the most relevant object in the image and verifying it through the entity selection, the fusion search process constructs object-aware textual queries by combining the user’s original question with the name of the verified object. This enables the system to issue more focused and semantically grounded web search queries, such as transforming a vague question like \textit{``What’s the price of this?''} into a more informative query like \textit{``Price of red sports car BMW M4.''}

\subsubsection{Text Search Engine}
To accommodate the retrieval needs of single or multiple sub-queries, we have built a custom text retrieval engine that can adapt to multiple sub-queries automatically. The engine performs query operations in the \textit{"crag-mm-2025/web-search-index-validation"}\cite{crag-mm-2025} index and uses the \textit{"bge-large-en-v1.5"} model\cite{bge_embedding} to calculate the similarity between the search results and the query's semantic embeddings. Finally, we fuse multiple retrieval results and reorder them based on similarity to obtain more relevant document content, thereby supporting subsequent responses.

\begin{figure}
    \centering
    \includegraphics[width=\linewidth]{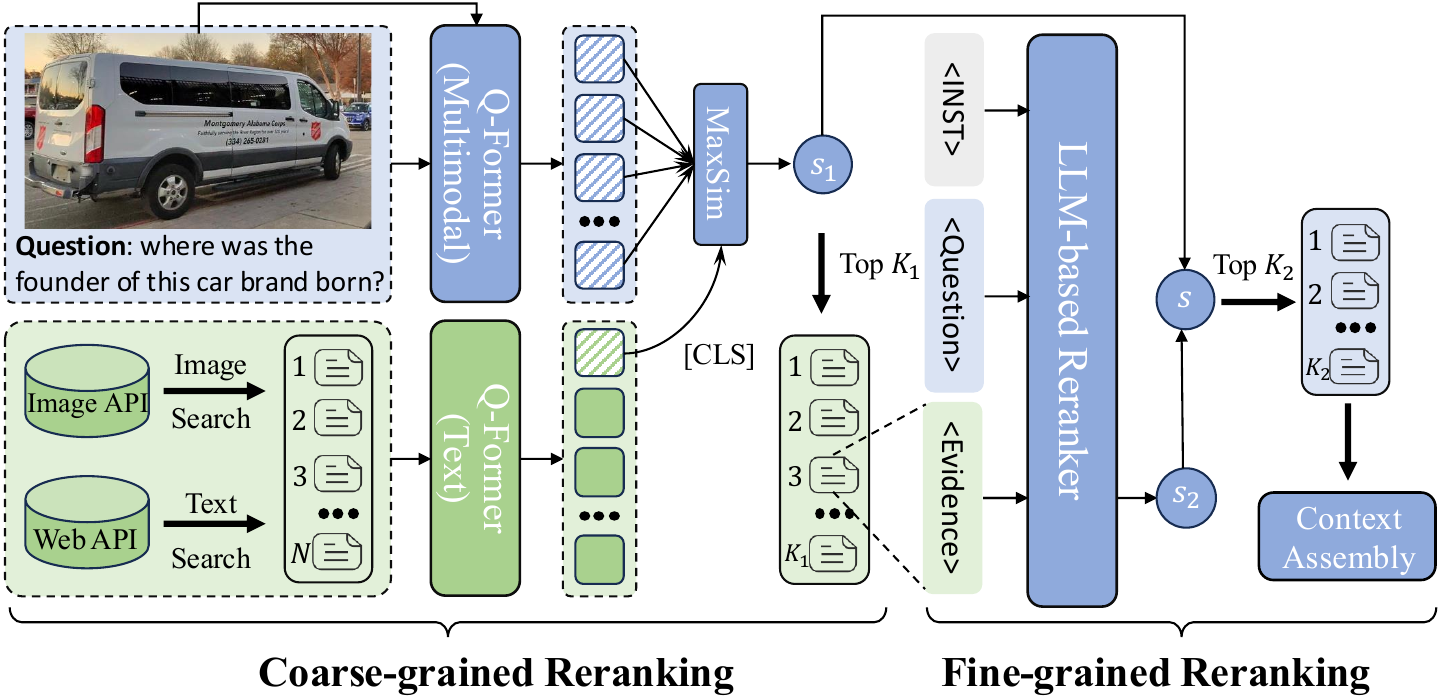}
    \caption{The pipeline of multimodal reranker.}
    \label{fig:reranker}
    \vspace{-0.5cm}
\end{figure}

\subsection{Coarse-to-fine Multimodal Reranker}\label{sec:reranker}
To efficiently use evidence retrieved from image and text search agents and filter query-relevant information, we employ a two-stage coarse-to-fine multimodal reranker that performs \textbf{Evidence Chunking}, \textbf{Multimodal Reranking}, and \textbf{Context Assembly}.

\subsubsection{Evidence Chunking.}
Let $\mathcal{D}_{\text{text}}$ and $\mathcal{D}_{\text{img}}$ denote the textual and image-based retrieval sets, respectively. The initial evidence set is defined as $\mathcal{D}_{\text{all}} = \mathcal{D}_{\text{text}} \cup \mathcal{D}_{\text{img}}$, where  $d \in \mathcal{D}_{\text{all}}$ is a paragraph from retrieved passages or an attribute block from image metadata. 
For each paragraph $d$, we first segment it into manageable pieces using a hierarchical chunking function $f_{\text{chunk}}(d)$, which takes into account the document structure (e.g., titles, paragraphs) as well as fixed-length spans. 
Given an attribute block, we construct a sentence for each attribute using the template: “The <key> of <entity> is <value>”, where <entity> refers to the entity name from the metadata. By concatenating these sentences, we form a paragraph corresponding to the attribute block. 
This generated paragraph is then processed using $f_{\text{chunk}}(d)$.
Finally, the chunked evidence set can be obtained by
\begin{equation}
\mathcal{C} = \bigcup_{d \in \mathcal{D}_{\text{all}}} f_{\text{chunk}}(d),
\end{equation}
where $c_i \in \mathcal{C}$ is a candidate evidence.

\subsubsection{Multimodal Reranking.}
We perform cascaded reranking for $\mathcal{C}$ in two stages, as depicted in Figure 3.

\textbf{Coarse-grained Reranking:} To ensure the relevance with both the question and image, a vision-language model $f_{\text{qformer}}$ based on \textsc{Q-Former}~\cite{li2023blip} is first used to filter evidence at a coarse granularity. 
Given a question-image pair $(q, x)$, \textsc{Q-Former} encodes it into a set of query tokens $\{z_q^{1},...,z_q^{N_q}\}$. On the candidate side, $c_i$ is encoded into a single \verb|[CLS]| token $z_t$ by \textsc{Q-Former}’s text encoder. Then, the relevance score for each evidence is computed via the following similarity function:
\begin{equation}
s^{(i)}_{\text{coarse}} = \max_{1\leq j \leq N_q} (\text{sim}(z_q^{j}, z_t)),
\end{equation}
which measures the maximum pairwise similarity between the query token embedding of $(q, x)$ and the [CLS] token embedding of $c_i$. 
Subsequently, the top $K_1$ candidates with the highest scores are selected. However, any candidate with scores below a predefined threshold $\tau_{\text{coarse}}$ is discarded.

\textbf{Fine-grained Reranking:} To more accurately evaluate similarity, we employ an LLM-based reranker (\textsc{Qwen3-Reranker}~\cite{zhang2025qwen3}) for point-wise reranking within a single context. 
Specifically, each evidence from the previous step is combined with the question and a well-designed instruction to form an input context, which is then fed into \textsc{Qwen3-Reranker} $f_{\text{qwen}}$ to compute a relevance score, \textit{i.e.},
\begin{equation}
s^{(i)}_{\text{fine}} = f_{\text{qwen}}(q, c_i | \text{instruction}).
\end{equation}
In this step, we select the top-$K_2$ chunks with the highest cumulative score, \textit{i.e.}, $s^{(i)} = s^{(i)}_{\text{fine}} \times s^{(i)}_{\text{coarse}}$. Similar to the first stage, only chunks with a score above $\tau_{\text{fine}} \times \tau_{\text{coarse}}$ are retained.

\subsubsection{Context Assembly.}
For each query, we reorganize the selected chunks into a final evidence string as:
\begin{equation}
\mathcal{E} = \texttt{Join} \left( \left\{ \texttt{Sort} ( c_i \right)\}_{i=1}^{K_2} \right),
\end{equation}
which serves as the input to the post answer generator and verifier.

This coarse-to-fine multimodal reranking mechanism filters noisy or redundant retrievals out while retaining semantically grounded, query-relevant evidence from both retrieval sources.

\subsection{Post Answer Generator \& Verifier}
Given the question-image pair ($q$, $x$), and the reranked evidence context $\mathcal{E}$ (as constructed in Section~\ref{sec:reranker}), we employ a CoT-based generator followed by an answer verifier to produce the final answer.

\subsubsection{CoT-based Answer Generation.}
For complex multimodal reasoning tasks, the capability of the MLLM remains limited, as demonstrated in Table~\ref{tab:main-ret}.
To enhance its reasoning performance, we employ CoT prompting techniques~\cite{wei2022chain} and ICL examples~\cite{dong2024survey}, guiding the model to articulate intermediate steps during problem-solving.
The MLLM is prompted to produce a detailed reasoning process and a concise final answer for each query.
The reasoning involves identifying supporting evidence from $\mathcal{E}$ relevant to the query, while the final answer addresses the query directly.
These intermediate reasoning steps also support answer verification in the subsequent stage. Refer to the Appendix for detailed templates.
 
\subsubsection{Answer Verifier.}
To reduce hallucinations and ensure factual precision, we implement a dual-verification mechanism to assess answer reliability.
First, we explore a lightweight white-box verifier that leverages the token probabilities of the generated response to quantify uncertainty~\cite{bouchard2025uncertainty}. 
Let $H=\{h_1, h_2, \dots, h_T\}$ denote the hidden states at the final decoder layer for each output token from the CoT-based Answer Generator.
We compute two statistical metrics: \textit{Minimum Token Probability} and \textit{Normalized Token Probability}, \emph{i.e.}, $s_{\text{min}} = \min_{i} t_i$ and $s_{\text{mean}} = \frac{1}{T} \sum_{i=1}^{T} t_i$.
Here, $t_i$ denotes the token probability, while $T$ is the number of tokens.
Then, both $s_{\text{min}}$ and $s_{\text{mean}}$ are passed to a linear thresholding function $g_d(v)$ that decides whether to accept or reject the generated answer. Only when $g_d(v) \geq \tau$ is the previous answer accepted as final. Otherwise, the system abstains from answering or falls back to a default statement (e.g., “I don't know”). 
Both $s_{\text{min}}$ and $s_{\text{mean}}$ are then fed into a linear thresholding function $g_{white}$, which determines whether to accept or reject the generated answer.
The answer is accepted only if $g_{white}(s_{\text{min}}, s_{\text{mean}}) \geq \tau_{white}$; otherwise, the system defaults to a fallback response (\emph{e.g.}, “I don’t know”).

\begin{table*}[h]
    \centering
    \caption{Overall Performance of Our Framework across Single-Source and Multi-Source Tasks.}
    \begin{tabular}{llllllcccccc}
        \toprule
        \multirow{2}{*}{\textbf{Method}} & \multicolumn{5}{c}{\textbf{Components}} & \multicolumn{3}{c}{\textbf{Single-Source}} & \multicolumn{3}{c}{\textbf{Multi-Source}} \\
        \cmidrule(lr){2-6} \cmidrule(lr){7-9}\cmidrule(lr){10-12}
        & DR & TR & FS & SR & FT & Accuracy ($\uparrow$) & Overlap ($\uparrow$) & Elapse ($\downarrow$) & Accuracy ($\uparrow$) & Overlap ($\uparrow$) & Elapse ($\downarrow$) \\
        \midrule
        LLM Only & -- & -- & -- & -- & --   & 15.79 & 18.98 & 0.71 & 15.79  & 18.98 & 0.71 \\
        CoT & -- & -- & -- & -- & --        & 16.25  & 20.07  & 1.78 & 16.25 & 20.07 & 1.78 \\
        Direct RAG & -- & -- & -- & -- & -- & 14.40 & 22.67 & 1.62 & 16.87 & 27.15 & 1.95  \\
        \midrule
        QA-Dragon$^\dag$ & \ding{51}& \ding{51}& \ding{55}& \ding{55}& \ding{55}  & 16.77 & 31.18 & 3.23 & 18.99 & 34.24 & 3.27  \\
        QA-Dragon$^\ddag$ & \ding{51}& \ding{51}& \ding{51}& \ding{55}& \ding{55}  & 17.44 & 33.57 & 3.32 & 22.08 & 42.62 & 3.83  \\
        QA-Dragon & \ding{51}& \ding{51}& \ding{51}& \ding{51}& \ding{55}  & 21.31 & 41.09 & 4.15 & 23.22 & 41.77 & 4.79  \\
        QA-Dragon$^{\star}$ & \ding{51}& \ding{51}& \ding{51}& \ding{51}& \ding{51} & 20.79 & 41.42 & 4.85 & 22.39 & 41.65 & 5.97  \\
        \bottomrule
    \end{tabular}
    \label{tab:main-ret}
    
    \vspace{0.1cm}\hspace{1cm}\raggedright
    \small{\textit{*\textbf{DR:} Domain Router; \textbf{TR:} Tool Router; \textbf{FS:} Fusion Search; \textbf{SR:} Search Router; \textbf{FT:} Finetuning.}}
\end{table*}

Moreover, we design an MLLM-based verifier to assess whether the reasoning produced by the CoT-based Answer Generator is logically sound and supported by the retrieved evidence.
Specifically, the MLLM is provided with the question-image pair $(q, x)$, the evidence context $\mathcal{E}$, and the reasoning response, and is tasked with classifying the reasoning as either ``Correct'' or ``Incorrect.'' 
The detailed prompt used for this step is provided in the Appendix.

Finally, we integrate the white-box and MLLM-based verifiers to yield high-certainty answers. This dual-verification approach ensures that final answers are grounded in the retrieved context and internally coherent and filtered by confidence, thereby reducing hallucinations in open-ended generation.

\begin{table}[t]
    \centering
    \caption{Overall Performance of Our Framework on Multi-Turn Task.}
    \label{tab:main-multi-turn-ret}
    \begin{tabular}{lccc}
        \toprule
        \textbf{Method}  & Accuracy ($\uparrow$) & Overlap ($\uparrow$) & Elapse ($\downarrow$)\\
        \midrule
        LLM Only   & 19.75 & 19.72 & 0.71\\
        CoT &  19.47 & 20.27 & 1.78\\
        Direct RAG & 20.95 & 33.74 & 1.95 \\
        \midrule
        QA-Dragon  & 24.78 & 48.26 & 7.00 \\
        QA-Dragon$^{\star}$ & 23.94 & 47.49 & 7.80 \\
        \bottomrule
    \end{tabular}
\end{table}

\section{Experiments}
In this section, we report our main results and conduct an ablation study to demonstrate the effectiveness of our key components.

\subsection{Experiments Setup}
According to the rules set by the organizers, we conduct our experiment on a single Nvidia L40s with 48GB of memory. The network connection is turned off during the inference process, meaning we can only leverage the knowledge from the given database. It also requires that each query be addressed within 10 seconds and labeled as failed after that.

In the model setup, we use \textit{Llama3.2-11B-Vision}\footnote{https://huggingface.co/meta-llama/Llama-3.2-11B-Vision} for all our MLLMs. The search API supports both text search and image search, which employs \textit{``BAAI/bge-large-en-v1.5''} and \textit{``openai/clip-vit-large-patch14-336''} for text and image embedding generation, respectively.

In our experiments, we compare our method with the following baselines. (1) \textbf{LLM Only} prompts the model to answer without retrieval; (2) \textbf{CoT} adds step-by-step reasoning. (3) \textbf{Direct RAG} retrieves references using the original query and image, and generates answers based solely on the retrieved content. Furthermore, during the development process, we conducted an in-depth analysis of the model outputs and rapidly iterated through four versions of the model:
\begin{itemize}
    \item \textbf{QA-Dragon$^\dag$:} Utilize the domain router and tool router to process the image search and text search.
    \item \textbf{QA-Dragon$^\ddag$:} Fuse the image search result into the text search during the search process.
    \item \textbf{QA-Dragon:} Propose a search router to categorize VQA queries into three search strategies.
    \item \textbf{QA-Dragon$^{\star}$:} Integrated a finetuning module to enhance the reasoning capability.
\end{itemize}

\subsubsection{Metrics}
In line with the MM-CRAG Benchmark, we conduct a model-based automatic evaluation for our experiment. The automatic evaluation employs rule-based \textbf{Overlap} matching and prompt \textit{GPT-4o-mini} as an assessment to check the \textbf{Accuracy} of the answer. Moreover, we report the average second \textbf{Elapse} of each query since efficiency is a critical aspect required by the organizers. The final score of each method is computed as the average score across all examples in the evaluation set.

\begin{table}[t]
    \centering
    \caption{Ablation Study for Key Components in Our QA-Dragon Framework.}
    \label{tab:ablation}
    \setlength{\tabcolsep}{3pt}
    \resizebox{\linewidth}{!}{
    \begin{tabular}{lcccccc}
        \toprule
        \multirow{2}{*}{\textbf{Method}} & \multicolumn{3}{c}{\textbf{Single-Source}} & \multicolumn{3}{c}{\textbf{Multi-Source}} \\
        \cmidrule(lr){2-4}\cmidrule(lr){5-7}
        & Acc. & Overlap & Elapse & Acc. & Overlap & Elapse \\
        \midrule
        QA-Dragon  & 21.31 & 41.09 & 4.15 & 23.22 & 41.77 & 4.79  \\
        \midrule
        \emph{w/o} Domain Router & 19.04 & 41.07 & 4.19 & 21.25 & 41.85 & 5.07 \\
        \emph{w/o} Tool Router & 18.32 & 41.49 & 4.02 & 19.97 & 42.32 & 4.25 \\
        \emph{w/o} Query Splitting & -- & -- & -- & 18.32 &  41.70 &  4.02 \\
        \emph{w/o} Reranking & 20.90 & 41.30 & 4.15 & 22.14 & 41.19 & 5.07 \\
        \bottomrule
    \end{tabular}}

    \vspace{0.1cm}\hspace{0.1cm}\raggedright \small{\textit{*Query Splitting is not applied under the single-source scenario.}}
\end{table}

\subsection{Overall Performance}
Table~\ref{tab:main-ret} and Table~\ref{tab:main-multi-turn-ret} report the performance of our method and baselines. Comparing our solutions to the LLM and RAG baselines, we observe significant advantages in performance across all three tasks. Our method showcases notable improvements in accuracy and information retention. Specifically: \textbf{QA-Dragon and QA-Dragon$^{\star}$ Achieves the Best Overall Performance}. QA-Dragon outperforms all baselines and ablations in both single-source (21.31\%), multi-source (23.22\%) and multi-turn (24.78\%) settings. The QA-Dragon$^{\star}$ achieves the highest overlap scores (41.42\% and 41.65\%), indicating better alignment with ground-truth supporting evidence. \textbf{Baselines Fall Short}. The LLM Only, CoT, and Direct RAG baselines significantly underperform in accuracy and overlap. \textbf{Trade-off between Performance and Efficiency}. The most powerful variant, QA-Dragon$^{\star}$, achieves the best performance but incurs the highest latency (5.97s).

\subsection{Ablation Study}
To assess the contribution of each core component in our framework, we conduct a series of ablation experiments. Table~\ref{tab:ablation} reports the results in terms of answer accuracy, entity overlap with ground truth, and response latency, to evaluate both correctness, efficiency, and alignment with human-annotated outputs.

\textbf{Domain Router.}  Removing the domain router leads to a noticeable drop in accuracy for both single-source (from 21.31\% to 19.04\%) and multi-source settings (from 23.22\% to 21.25\%). It demonstrates the importance of domain-aware query routing in selecting appropriate retrieval strategies, especially when grounding queries in specific content domains.

\textbf{Tool Router.} Disabling the tool router results in further performance degradation, particularly in the single-source setting (from 21.31\% to 18.32\%). Interestingly, overlap remains high (41.49\% and 42.32\%), suggesting that while relevant content is still retrieved, the absence of precise tool selection limits the model’s ability to convert evidence into correct answers.

\textbf{Query Splitting.} 
Eliminating query splitting, which decomposes complex questions into interpretable sub-goals, significantly degrades the performance. This result validates the effectiveness of breaking down complex queries into multiple logical sub-queries to enable more rigorous textual retrieval.

\textbf{Two-stage Reranking.}  We further test the effect of removing two-stage reranking in the retrieval pipeline. Accuracy slightly decreases (from 21.31\% to 20.90\% in single-source and from 23.22\% to 22.14\% in multi-source), and latency increases by nearly 0.3s in the multi-source case. This indicates that reranking retrieved candidates meaningfully improves answer selection without compromising efficiency.

These results highlight that each component in our system contributes meaningfully to either performance or latency. The framework achieves the best overall balance by dynamically coordinating reasoning and retrieval.

\begin{figure}
    \centering
    \includegraphics[width=\linewidth]{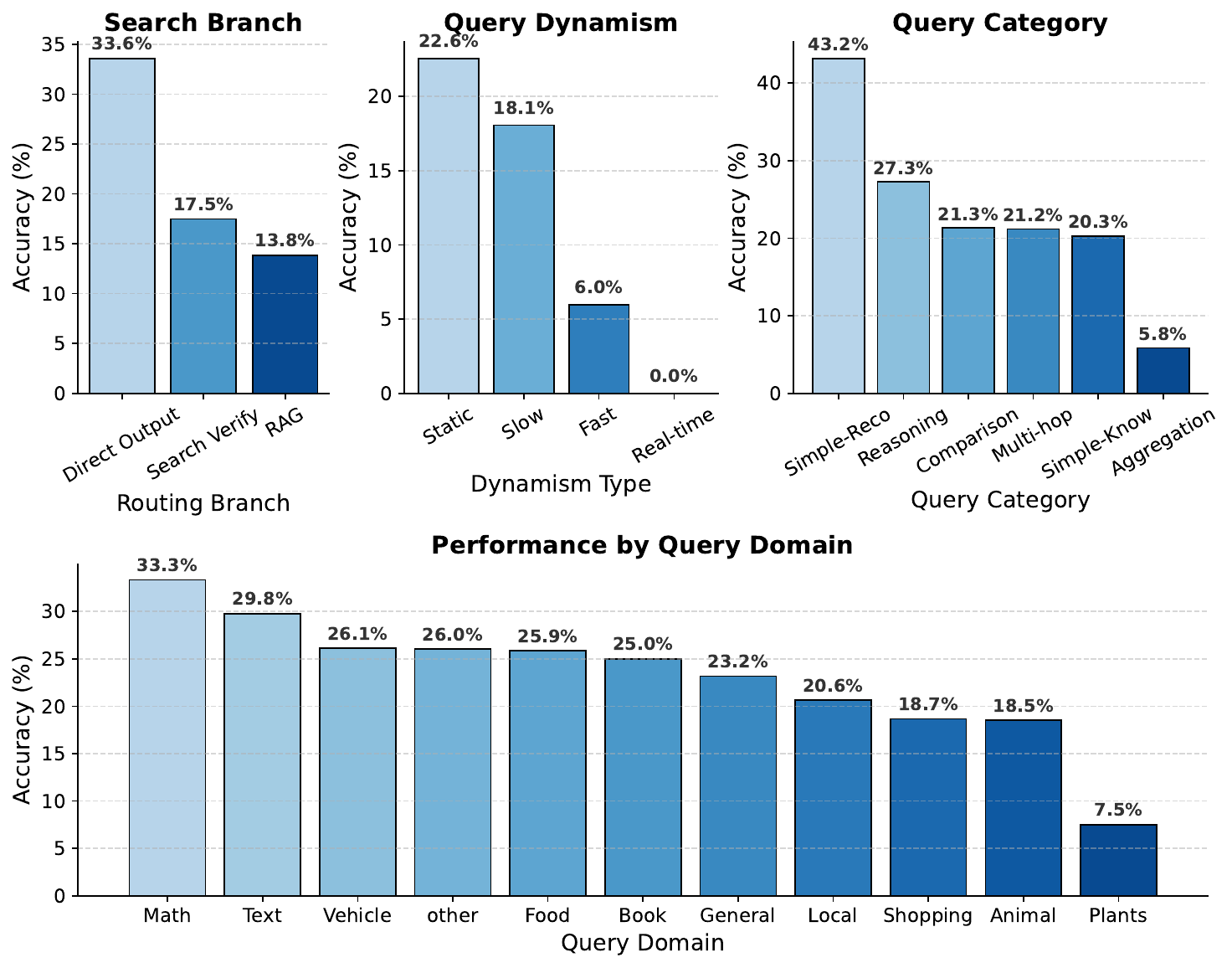}
    \caption{Accuracy over different query taxonomies on the single-source task.}
    \label{fig:single-source}
\end{figure}

\begin{figure}
    \centering
    \includegraphics[width=\linewidth]{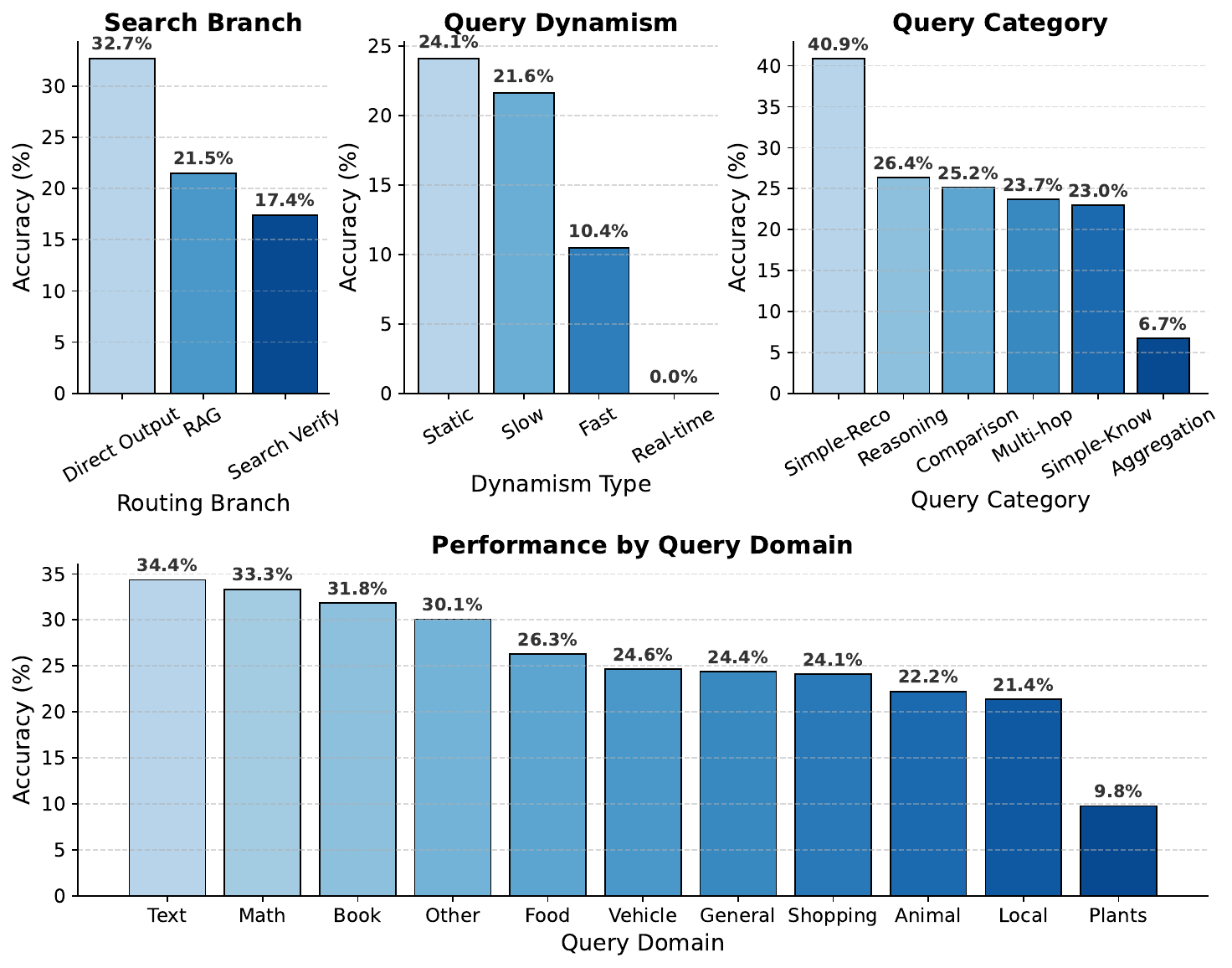}
    \caption{Accuracy over different query taxonomies on the multi-source task.}
    \label{fig:multi-source}
\end{figure}

\section{Perspectives}
To better understand the behavior and limitations of our QA-Dragon framework, we conduct a detailed evaluation across four query taxonomies: search branch, query dynamism, query category, and query domain, as shown in Figure~\ref{fig:single-source} and~\ref{fig:multi-source}, respectively.

Across both tasks, we observe several consistent patterns. Queries routed to the \textit{Direct Output} branch achieve the highest accuracy (33.6\% for single-source and 32.7\% for multi-source), significantly outperforming \textit{Search Verify} and \textit{RAG Augment}, highlighting the effectiveness of the \textit{Search Router} in identifying queries that can be resolved through image-grounded or self-contained reasoning directly. Accuracy decreases steadily as \textit{query dynamism} increases, with real-time queries yielding no correct answers. This illustrates the challenge of temporally sensitive reasoning and emphasizes the need for improved temporal grounding in future systems. In terms of query category, \textit{Simple Recognition} achieves the highest performance (43.2\% and 40.9\%), while \textit{Aggregation} is the most difficult (5.8\% and 6.7\%), likely due to its multi-hop and cross-source nature. Among query domains, \textit{Math} and \textit{Text} lead in accuracy, benefiting from structured formats and OCR-friendly content. Conversely, the \textit{Plant} domain shows the lowest performance (7.5\% and 9.8\%), reflecting the difficulty of fine-grained visual discrimination in biologically similar classes.

\section{Conclusion}

In this work, we proposed the \textbf{QA-Dragon}, a query-aware dynamic RAG system tailored for complex and knowledge-intensive VQA tasks. Unlike the traditional RAG system, which retrieves the text and image separately, the QA-Dragon integrates a domain router, a search router, and orchestrates hybrid retrieval strategies via dedicated text and image search agents.  
Our system supports multimodal, multi-turn, and multi-hop reasoning through orchestrating both text and image search agents in a hybrid setup. Extensive evaluations on the Meta CRAG-MM Challenge (KDD Cup 2025) demonstrate that our system significantly improves answer accuracy, surpassing other strong baselines.

\clearpage

\balance
\bibliographystyle{ACM-Reference-Format}
\bibliography{bibliography}

\onecolumn
\appendix
\section{Prompts}
\label{sec:prompt}
\newtcolorbox{mybox}[2][]{%
  breakable,
  enhanced,
  width=\linewidth, 
  boxrule=0.5pt,     
  arc=2mm,           
  colback=gray!10,   
  colframe=gray!50,  
  coltitle=white,    
  colbacktitle=gray!70, 
  fonttitle=\bfseries,
  title=#2,
  #1
}

\begin{mybox}{Evaluator Prompt}
\textbf{System Prompt:}

You are a visual assistant tasked with addressing the user's query for the image based on your inherent knowledge.

\textbf{General Reasoning Guidelines:}
    \begin{enumerate}
        \item Generate step-by-step reasoning to address the query using evidence from the image and your knowledge. Limit your reasoning to no more than 5 concise steps, with each step written as a single sentence. Stop reasoning once you have enough information to answer the query or when you determine that necessary information is lacking.
        \item In your reasoning, identify the exact object that the query is about by its exact name (\emph{e.g.}, car model, food name, brand name, species name, etc.). If no clear object matches the query, you may refer to textual clues visible in the image if available.
        \item If the query involves multiple objects or relationships, dedicate one reasoning step to each object or relationship, and then summarize the result in a final step.
        For example:
        \begin{enumerate}
            \item The exact name of the object in the image that the query is about is <specific\_object\_name>.
            \item Next, the exact name of the object related to the first one is <specific\_object\_name>."
            \item ...
        \end{enumerate}
        \item If you cannot determine the necessary information from the image or the query, explicitly state: "I cannot determine the <what> that the query is about."
        \item Do not suggest that the user to refer to external sources.
        \item Always begin your reasoning with: "1. The exact name of the object that the query "<query>" is about is <specific\_object\_name>." and make your final reasoning concise.
    \end{enumerate}

    \verb|<Doamin-aware ICL cases>|
    
\textbf{Output Format:}
    \begin{enumerate}
        \item The exact name of the object in the image that the query is about is <specific\_object\_name>.
        \item Then, I ...
        \item ...
    \end{enumerate}
    \verb|{"reasoning": "<summary_reasoning_string>"}|
\end{mybox}

\begin{mybox}{Image Object List Prompt}
\textbf{System Prompt:}

You are an expert AI system for object detection and identification. Your task is to recognize and list high-level object categories shown in an image that are relevant to a given question. Return only structured results in JSON format.

\textbf{User Prompt:}

Identify and list up to \verb|{self.object_num}| major distinct objects in the image that are visually present and relevant to the question: "{query}". 

Only include tangible, visible items (e.g., "car", "brand", "clothing", "book", "device", "food", "building"). 

Do not include abstract concepts (e.g., "emotion", "relationship") or actions (e.g., "running", "shopping").

Use general categories, not specific names: "BMW" to "car", "ZARA" to "clothing brand", "iPhone 13" to "smartphone", "Coca-Cola" to "drink"

Each object name should be short with no more than 3 words.

If unsure of the exact identity, use the closest general category (e.g., "electronic device", "building", "plant").

Return the result in the following JSON format strictly:

\verb|{"object_list": ["<object_name_string>", "<object_name_string>", ...]}|
\end{mybox}

\begin{mybox}{Image Object Selection Prompt}
\textbf{System Prompt:}
You are an AI assistant to select one object from a list of objects, which is most relevant to the object queried by the question. Only return structured results in JSON format.

\textbf{User Prompt:}
Given the image and list of objects detected in the image: \verb|{object_list}| and the question: \verb|"{query}"| select the one object in the object list that the question is about.

If the query includes position-related words, give priority to objects at or near the position.

Give a short sentence about the reason to choose the object and return the final selected object in this format:

\verb|{"object": "<object_name_string>"}|
\end{mybox}

\begin{mybox}{Evaluator Prompt}
\textbf{System Prompt:}
You are an action-selector. Given three inputs: (1) the user's query, (2) any prior reasoning text, and (3) the image, decide in a single turn which retrieval tool or tools must run before the answer is generated. You may choose image\_search, text\_search, both, or neither.

\textbf{Tools:}
    \textbf{Tool \#1 image\_search}
    
    [Description] Retrieve visually similar images via embeddings to identify an object whose specific name is still unknown.
    
    [Use when] The object in the picture is not known or known only by a generic label (e.g., “car”, “jacket”, “statue”) instead of a specific name/model/species in previous reasoning.
    
    [Input] The input image.
    
    [Output] Text snapshots (top-k) from Wikipedia or Amazon that show visually similar objects.
    
    \textbf{Tool \#2 text\_search}
    
    [Description] Issue a refined natural-language web query to fetch textual facts about an object whose specific name is already known.
    
    [Use when] The query requires additional information not available in the image.
    
    [Input] A text query constructed from the user's question + the known identity.
    
    [Output] Text snippets (top-k) from relevant websites.
    
\textbf{Decision logic}
    \begin{enumerate}
        \item Do you or the reasoning text already know the object's specific identity (proper noun or model name)?
        
        - Yes: set `need\_image\_search` to false.
        
        - No: set `need\_image\_search` to true.
        
        \item Does the query need additional information that is not visible in the image (specifications, history, statistics, price, etc.)?
        
        - Yes: set `need\_text\_search` to true.
        
        - No: set `need\_text\_search` to false.
        
        \item If it is about addressing some scientific calculation queries like math, physics, etc., or language translation, set both flags to false.
        \item If the object is a "book", a "logo-bearing packaged goods", or "plant", set `need\_image\_search` to false.
    Both flags may be true; when so, run image\_search first, then text\_search.
    \end{enumerate}
Produce exactly one sentence to conduct the decision logic. This is the only non-JSON text allowed.

Immediately after the sentence, output a single valid JSON object:
Decision logic: <concise explanation>

Tool calling decision: {{"need\_image\_search": <true/false>, "need\_text\_search": <true/false>}}

Do not output anything else.

Here are some examples:

    \verb|<Tool Using ICL cases>|
    
\end{mybox}

\begin{mybox}{Post Answer Generation Prompt}
\textbf{System Prompt:}
You are a helpful assistant who truthfully answers user questions about the provided image. If you are not sure about the answer, please say 'I don't know.'.

\textbf{User Prompt:}
Answer the given question based on the provided image and your own knowledge. Please think step by step and give a response containing the following parts:
\begin{itemize}
    \item reason: the information from the image or your own knowledge that leads to the answer, which should be clear and concise within 2-3 sentences. If you are not sure about the answer, please say 'I don't know.'.
    \item answer: your final answer in a concise format within one sentence. The answer should include critical information from the reason to support your answer. When referring to a specific object in the image, please use the name of the object, rather than 'this', 'that', or 'it'. If the reason is 'I don't know, ' please also say 'I don't know' in the answer.
\end{itemize}


-Examples-\\
\verb|<ICL examples>|

-Real Data-\\
Question: \verb|<question>|\\

Your Output:
\end{mybox}

\begin{mybox}{Answer Verifier Prompt}
\textbf{System Prompt:}
You are a helpful assistant who evaluates whether the agent's answer to the user's image query is reasonable based on the evidence.

\textbf{User Prompt:}
Given an image query, the retrieved evidence, and the agent’s candidate answer, assess the correctness of the answer by following these guidelines:
\begin{enumerate}
\item Unsupported Answer. If the answer is not supported by the image and evidence, please respond the follows:\\
**Reason:** Briefly (1–2 sentences) explain why the answer lacks sufficient support.\\
**Response:** Incorrect Answer
\item Contradicted Evidence. If there is conflicting information in the evidence, please respond the follows:\\
**Reason:** Briefly (1–2 sentences) state the specific contradictory evidence.\\
**Response:** Incorrect Answer
\item  Unclear or Incomplete Answer. If the answer is vague or fails to fully address the question, please respond the follows:\\
**Reason:** Briefly (1–2 sentences) explain why the answer is unclear or incomplete.\\
**Response:** Incorrect Answer
\item Correct Answer. If the answer is fully supported by the image and evidence, please respond the follows:\\
**Reason:** Briefly (1–2 sentences) state why the answer is correct.\\
**Response:** Correct Answer
\end{enumerate}


-Real Input-\\
Query: \verb|<quesion>|\\
Evidence: \verb|<evidence>|\\
Answer: \verb|<answer>|\\

Your Output:
\end{mybox}

\end{document}